\documentclass[11pt]{article}

\PassOptionsToPackage{table}{xcolor}
\usepackage{acl}

\usepackage{times}
\usepackage{latexsym}
\usepackage{booktabs}
\usepackage{multirow}
\usepackage{amsmath}
\usepackage{amssymb}
\usepackage{amsfonts}
\usepackage{xcolor}
\usepackage{comment}
\usepackage{colortbl}
\usepackage[most]{tcolorbox}
\usepackage[T1]{fontenc}
\usepackage[utf8]{inputenc}
\usepackage{microtype}
\usepackage{algorithm}
\usepackage{algorithmic}
\usepackage{inconsolata}
\usepackage{graphicx}
\newcommand{\methodname}{\textsc{Siri}}

\title{\methodname: \underline{S}elf-\underline{I}nternalizing \underline{R}einforcement Learning with \underline{I}ntrinsic Skills for LLM Agent Training}

\author{
 \textbf{Zhongyu He\textsuperscript{1,2,*}},
 \textbf{Yuanfan Li\textsuperscript{2,*}},
 \textbf{Fei Huang\textsuperscript{2}},
 \textbf{Tianyu Chen\textsuperscript{2}},
 \textbf{Siyuan Chen\textsuperscript{2,\textdagger}},
\\
 \textbf{Xingyang Li\textsuperscript{2,\textdagger}},
 \textbf{Meng Hsuan Yu\textsuperscript{2}},
 \textbf{Xiangrong Liu \textsuperscript{1}},
 \textbf{Leyi Wei\textsuperscript{3}},
\\
 \textbf{Lu Pan\textsuperscript{2}},
 \textbf{Ke Zeng\textsuperscript{2}},
 \textbf{Xunliang Cai\textsuperscript{2}}
\\
 \textsuperscript{1}School of Informatics, Xiamen University~
 \textsuperscript{2}Meituan~
 \textsuperscript{3}Macao Polytechnic University
\\
 \texttt{hezhongyu@stu.xmu.edu.cn},  \texttt{ liyuan7716@gmail.com}\\
 \texttt{\{lixingyang04, chensiyuan25\}@meituan.com}\\
 \url{https://hezy618.github.io/SIRI}
}

\begin{document}
\maketitle
\begin{abstract}
Long-horizon LLM agents can benefit from reusable skills, yet existing skill-based methods often rely on external skill generators during training or persistent skill retrieval at inference, increasing engineering complexity, context length, and deployment latency. We propose \textbf{S}elf-\textbf{I}nternalizing \textbf{R}einforcement learning with \textbf{I}ntrinsic skills (\methodname{}), a three-phase framework that enables agents to discover, validate, and internalize skills without external skill generators or inference-time skill banks. \methodname{} first warms up the policy with GiGPO to acquire basic interaction ability and collect successful skill-free trajectories. It then performs self-skill mining, where the current policy summarizes compact skills from its own successful plain rollouts and validates them through paired skill-augmented and skill-free rollouts. Finally, \methodname{} distills only beneficial skill-guided action tokens into the plain policy using trajectory-level utility and action-level advantage. At inference, the agent runs with the original prompt only. On ALFWorld and WebShop with Qwen2.5-7B-Instruct, \methodname{} improves GiGPO from 0.908 to 0.930 on ALFWorld and from 0.728 to 0.813 on WebShop, outperforming prompt-based, RL-based, and memory-augmented baselines. Further analysis shows that our self-mining strategy can achieve performance comparable to distillation with closed-source large model.
\end{abstract}

\section{Introduction}
\begingroup
\renewcommand{\thefootnote}{*}
\footnotetext{These authors contributed equally to this work.}
\renewcommand{\thefootnote}{\dag}
\footnotetext{Corresponding author.}
\endgroup
\label{sec:intro}
\begin{figure}[t]
  \includegraphics[width=\columnwidth]{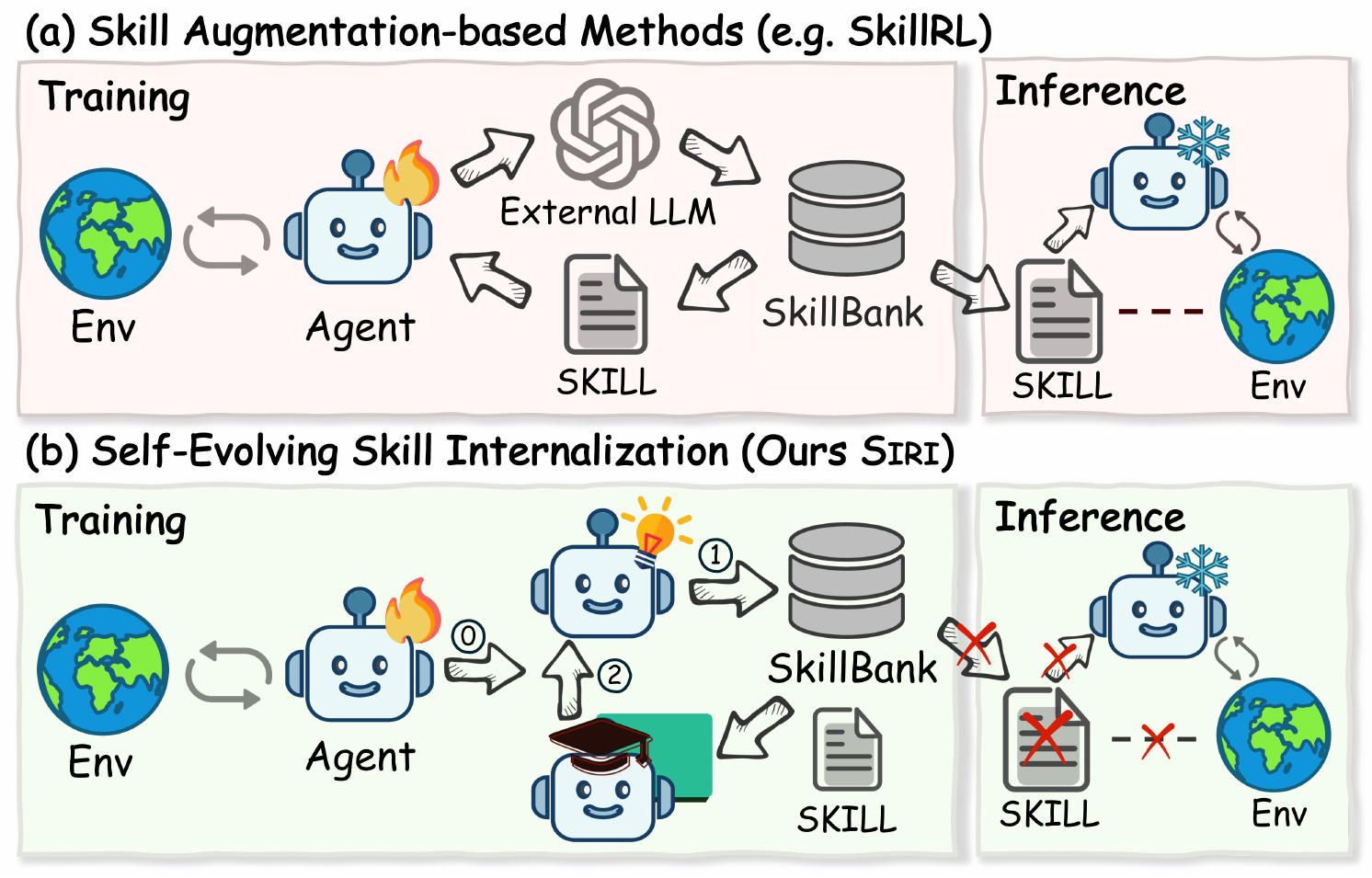}
    \caption{
        Conceptual comparison between (a) traditional skill-augmentation frameworks and (b) our \methodname{}.
    }
    \label{fig:motivation}
\end{figure}

Large language models (LLMs) have rapidly evolved from text generators into autonomous agents capable of interacting with external environments, using tools, and making long-horizon decisions~\cite{brown2020language,bai2022training,yao2023react}. 
Unlike standard single-turn generation tasks, agentic LLMs must execute sequences of environment-dependent actions, recover from intermediate mistakes, and optimize for delayed task success. 
This setting has motivated a growing body of reinforcement learning (RL) research for LLM agents in embodied household environments, web navigation, and application-centered tool execution~\cite{shridhar2021alfworld,yao2022webshop,trivedi2024appworld}. 
However, long-horizon agent learning remains challenging: sparse terminal rewards provide limited supervision for intermediate decisions, while repeatedly exploring large action spaces can be inefficient and unstable.

A promising direction is to equip agents with reusable \emph{skills}. 
Skills compress successful experience into high-level behavioral abstractions, enabling agents to reuse prior knowledge instead of rediscovering similar strategies from scratch. 
Early language agents store verbal reflections or episodic memories for later retrieval~\cite{shinn2023reflexion,zhao2024expel}, and recent skill-based RL methods further integrate skill banks into agent training~\cite{xia2026skillrl,tu2026d2skill,wang2026skillsd}. 
In long-horizon environments, such skills can provide useful guidance: for instance, a household agent may learn that when a target object is visible and reachable, it should pick it up before continuing exploration; a web-shopping agent may learn to first filter candidates by hard constraints before comparing soft preferences. 
These abstractions can reduce exploration burden and improve policy learning.

Despite their effectiveness, existing skill-based agent methods expose two important limitations, as illustrated in Figure~\ref{fig:motivation}. 
First, trajectory-to-skill conversion is often delegated to external or separately maintained skill-construction modules, such as trajectory distillers, skill-conditioned teachers, or hierarchical skill managers~\cite{xia2026skillrl,tu2026d2skill,wang2026skillsd,li2026arise}. 
This introduces additional engineering complexity, increases training cost, and makes the agent's improvement partially dependent on auxiliary models or modules rather than its own policy learning. 
Second, many skill-memory methods treat skills as persistent runtime context: the agent must retrieve relevant skills from a skill bank and insert them into the prompt during inference~\cite{shinn2023reflexion,zhao2024expel,wang2023voyager,xia2026skillrl,tu2026d2skill}. 
Although such retrieval can improve decision quality, it increases context length, retrieval latency, and deployment complexity, especially in long-horizon tasks where the agent repeatedly interacts with the environment.

These limitations raise a natural question: 
\emph{Can an agent discover useful skills by itself during RL training, and then absorb those skills into its own parameters so that no skill bank is needed at inference time?}
A positive answer requires solving two coupled problems. 
The first is \emph{skill discovery}: since the model is trained to act rather than to summarize, self-generated skills may be noisy or spurious if extracted from low-quality trajectories. 
The second is \emph{skill internalization}: even if retrieved skills improve skill-conditioned rollouts, directly imitating all skill-augmented responses may distill irrelevant language patterns rather than useful actions. 
A reliable framework must therefore determine both \emph{which skills are useful} and \emph{which skill-guided actions should be absorbed}.

To this end, we propose \textbf{S}elf-\textbf{I}nternalizing \textbf{R}einforcement learning with \textbf{I}ntrinsic skills (\textbf{\methodname{}}), a three-phase curriculum framework for autonomous skill discovery and skill-free policy learning in long-horizon agentic RL.
\methodname{} is designed around a central principle: RL credit should not only optimize actions, but also determine which skills are useful and which skill-guided behaviors should be absorbed into the policy.
In the first phase, \methodname{} performs policy warmup to bootstrap basic interaction ability and collect successful trajectories from the skill-free policy.
This phase provides high-quality behavioral evidence for subsequent skill extraction and avoids mining spurious skills from an undertrained agent.
In the second phase, \methodname{} conducts \emph{Self-Skill Mining and Utilization}, where the current policy itself summarizes compact skills from its own successful plain rollouts.
These self-generated skills are treated as candidates rather than trusted supervision: they are validated through paired skill-augmented and skill-free rollouts, and are promoted only when they yield positive online utility over the plain baseline.
This design enables skill discovery without relying on external skill generators.
In the third phase, \methodname{} performs \emph{Advantage-Weighted Skill Internalization}, which transfers useful skill-guided behaviors into the skill-free policy.
Specifically, trajectory-level treatment effects estimate whether retrieved skills improve the whole rollout, while action-level advantages identify which skill-guided action tokens are worth distilling.
Only action tokens from beneficial skill-conditioned trajectories contribute to the internalization loss.
As a result, skills serve as temporary training-time signals rather than permanent inference-time dependencies.
At deployment, the final policy uses the original prompt only, without any skill bank, retrieval service, or external memory.

We evaluate \methodname{} on ALFWorld~\cite{shridhar2021alfworld} and WebShop~\cite{yao2022webshop} with Qwen2.5-7B-Instruct. 
\methodname{} achieves the best overall performance among prompt-based agents, standard RL baselines, and memory-augmented RL methods. 
It improves GiGPO from 0.908 to \textbf{0.930} on ALFWorld, and from 0.728 to \textbf{0.813} success rate on WebShop, with the WebShop score increasing from 0.844 to \textbf{0.899}. 
Compared with SkillRL, \methodname{} improves ALFWorld success from 0.899 to \textbf{0.930} and WebShop success from 0.727 to \textbf{0.813}. 
We further replace the self-mining skill generator with stronger external LLM, and observe that the performance gap consistently narrows during training. 
This suggests that self-mined skills become increasingly effective through online utility validation and skill internalization.
Our contributions are summarized as follows:
\begin{itemize}
    \item \textbf{Self-mining skill learning framework.}
    We introduce a self-mining training framework that enables the agent to discover reusable skills from its own successful plain rollouts during reinforcement learning, without relying on external skill generators.

    \item \textbf{Retrieval-Free Inference via Advantage-weighted skill internalization.}
    We propose an advantage-weighted internalization mechanism that uses skills only as temporary training-time guidance and absorbs useful skill-guided actions into the policy, removing the need for an external skill bank at inference time.

    \item \textbf{Strong long-horizon agent performance.}
    Experiments on ALFWorld and WebShop show that \methodname{} achieves the best overall results, improving ALFWorld success from 0.908 to \textbf{0.930} and WebShop success from 0.728 to \textbf{0.813} over GiGPO, demonstrating effective skill internalization without inference-time retrieval.
\end{itemize}

\section{Related Work}

\noindent \textbf{Skills in reinforcement learning for LLM agents.}
Skills offer a compact abstraction for reusing experience in long-horizon agent learning. 
Early language agents store verbal reflections, episodic summaries, or non-parametric memories for later retrieval~\cite{shinn2023reflexion,zhao2024expel,mallen2023not}, while tool- and web-augmented agents further exploit external knowledge or executable behaviors to improve decision making~\cite{nakano2021webgpt,schick2023toolformer,wang2023voyager}. 
Recent reinforcement learning methods move skill acquisition into training. 
SkillRL distills trajectories into a hierarchical skill bank~\cite{xia2026skillrl}, while D2Skill organizes experience into task-level and step-level skills with utility-aware retrieval and pruning~\cite{tu2026d2skill}. 
Skill-SD uses skill-conditioned contexts as privileged guidance for self-distilling a plain-prompt policy~\cite{wang2026skillsd}, while Skill0 progressively withdraws in-context skill guidance through a helpfulness-driven dynamic curriculum to internalize skills into policy parameters~\cite{lu2026skill0}.
ARISE explores intrinsic skill evolution in hierarchical reinforcement learning for reasoning tasks~\cite{li2026arise}. 
Despite their effectiveness, these methods typically rely on skills as persistent runtime memory, and even when skills are internalized, they are either externally constructed or distilled wholesale without validating their actual utility.
In contrast, \methodname{} self-generates skills from its own successful plain rollouts, validates them through online paired-rollout utility, and selectively internalizes beneficial skill-guided actions into a policy that requires no skill bank or retrieval at inference time.

\noindent \textbf{Reinforcement learning for LLM agents.}
Reinforcement learning has become a standard approach for training LLM agents in interactive environments, from text games and language-action spaces to ALFWorld, WebShop, and AppWorld~\cite{mnih2015human,narasimhan2015language,he2016deep,hausknecht2020interactive,shridhar2021alfworld,yao2022webshop,trivedi2024appworld}. 
While prompt-based agents such as ReAct interleave reasoning and acting~\cite{yao2023react}, they do not directly learn from environment feedback. 
Recent methods therefore focus on long-horizon policy optimization and credit assignment~\cite{sutton2018reinforcement,pignatelli2023survey,dulac2015deep,zhang2026reasoning}: RAGEN studies self-evolution in multi-turn reinforcement learning~\cite{wang2025ragen}, Tree-GRPO derives process signals from tree-structured rollouts~\cite{ji2025tree}, and GiGPO estimates both episode-level and anchor-state step-level advantages from grouped rollouts~\cite{feng2025group}. 
Other work explores hierarchical training, online tuning, end-to-end web-agent RL, hindsight credit, and retrieval-augmented exploration~\cite{zhou2024archer,putta2024agent,feng2025towards,chen2025reinforcement,wei2025webagent,tan2026hindsight,zhang2026rapo}. 
Different from methods that only use credit for policy optimization, \methodname{} reuses GiGPO's two-level credit to validate generated skills and selectively internalize useful skill-guided actions into a no-skill inference policy.

\section{Preliminaries}
\label{sec:preliminaries}

\subsection{History-Augmented Decision Process}
We formulate the agent's long-horizon interaction as a Markov Decision Process (MDP) $\mathcal{M} = (\mathcal{S}, \mathcal{A}, P, r, \gamma)$. However, unlike classical reinforcement learning, LLM policies do not directly observe the underlying environment state $s_t \in \mathcal{S}$. Instead, they interact via a textual interface. For a given task $g$, let $I_\text{task}$ denote the base task instruction, $o_t$ the textual observation at step $t$, and $\mathcal{H}_t = \{(o_1, a_1), \dots, (o_{t-1}, a_{t-1})\}$ the accumulated interaction history. The policy generates actions $a_t \sim \pi_\theta(\cdot \mid x_t)$ conditioned on the effective context $x_t = (I_\text{task}, \mathcal{H}_t, o_t)$. Because $x_t$ is merely a summary of past interactions and generally not a sufficient statistic of the latent state, this process is fundamentally a history-augmented partially observable decision process.

\subsection{Skills for LLM Agents}

\begin{figure*}[t]
  \includegraphics[width=\linewidth]{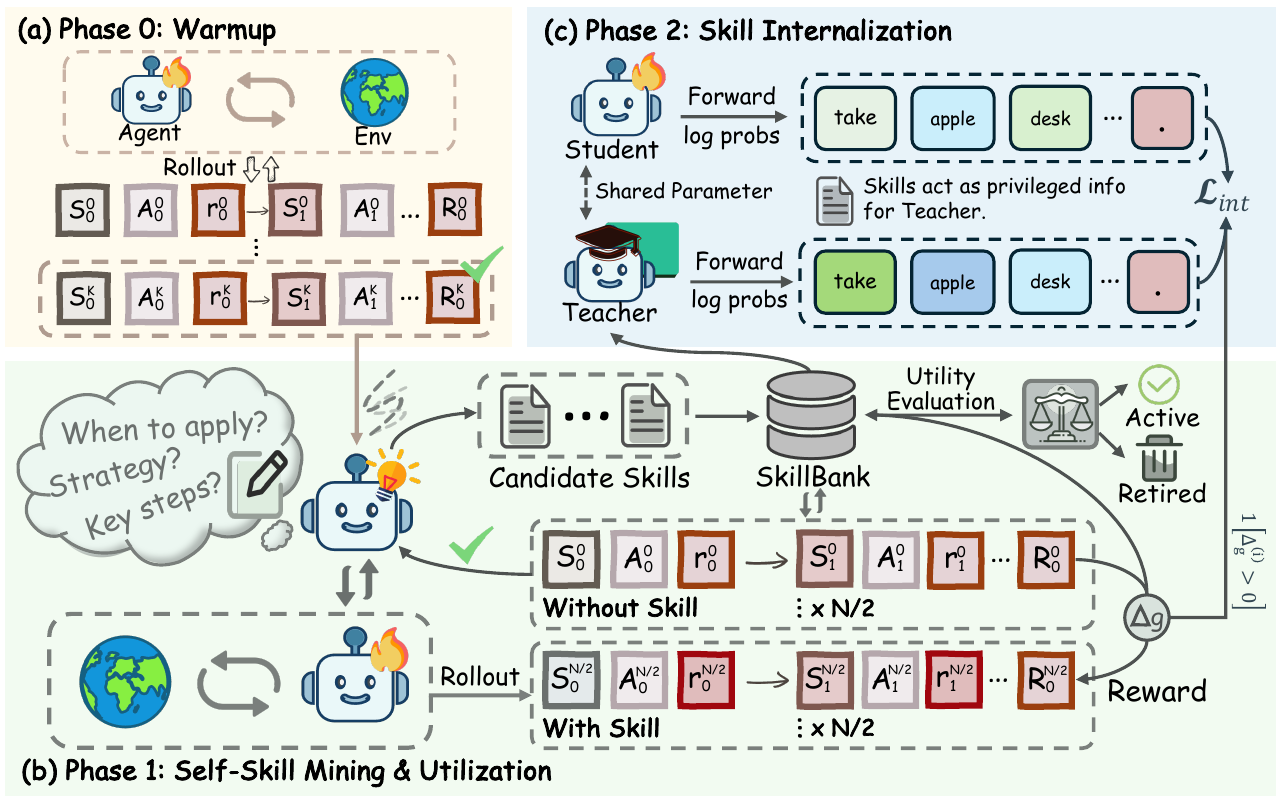}
  \centering
  \caption{\textbf{Overview of \methodname{}.} 
    \textbf{(a) Phase 0:} The agent interacts with the environment to collect initial trajectory data for cold-start policy initialization. 
    \textbf{(b) Phase 1:} Candidate skills are extracted into a structured \texttt{SkillBank}, followed by parallel rollouts with and without skills to compute the hindsight macro utility $\Delta_g$ for dynamic skill evaluation. 
    \textbf{(c) Phase 2:} The shared policy parameterizes validated skills via cross-context self-distillation under asymmetric inputs, optimized by an advantage-weighted loss $\mathcal{L}_{\text{int}}$ gated by the macro performance indicator $\mathbf{1}[\Delta_g^{(i)} > 0]$.}
    \label{fig:overall_framework}
\end{figure*}
In agentic reinforcement learning, a skill $m \in \mathcal{B}$ consists of a natural-language strategy $m.\text{strategy}$ that describes how to act and a scenario description $m.\text{condition}$ that characterizes when the skill applies. The agent maintains the dynamic skill library $\mathcal{B} = \{m_1, m_2, \dots\}$ as it continuously explores the environment. To reuse a skill, the agent retrieves relevant skills based on the current context and generates its action conditioned on both the task observation and the skill strategy:
\begin{equation}
a_t \sim \pi_\theta(\cdot \mid x_t, m.\text{strategy})
\end{equation}
This formulation effectively utilizes prior knowledge to guide exploration and decision-making in complex environments without requiring dense step-by-step external supervision.

\subsection{Group-in-Group Policy Optimization (GiGPO)}
GiGPO \citep{feng2025group} enables fine-grained credit assignment in long-horizon agent training without an external critic. Given a group $\mathcal{G}$ of $N$ trajectories sampled from the same task, it computes advantages at two granularities.

\noindent \textbf{Episode-Level Advantage.} The return $R(\tau_i)$ of each trajectory is normalized within $\mathcal{G}$:
\begin{equation}
A^E(\tau_i) = \frac{R(\tau_i) - \mu_{\mathcal{G}}}{\sigma_{\mathcal{G}}}
\end{equation}

\noindent \textbf{Step-Level Advantage.} GiGPO identifies \textit{anchor states}, environment states revisited across trajectories, and groups the actions taken from each anchor state $\tilde{s}$ into a step-level group $G^S(\tilde{s})$. The discounted future return $R_t^{(i)} = \sum_{k=t}^{H} \gamma^{k-t} r_k^{(i)}$ is then normalized within this group:
\begin{equation}
A^S(a_t^{(i)}) = \frac{R_t^{(i)} - \mu_{G^S}}{\sigma_{G^S}}
\end{equation}

The composite advantage $A(a_t^{(i)}) = A^E(\tau_i) + \omega \cdot A^S(a_t^{(i)})$ fuses trajectory-level and step-level signals, and is used in the standard clipped surrogate objective:
\begin{equation}
\begin{split}
\mathcal{L}_{\text{GiGPO}}(\theta) = &-\frac{1}{N} \sum_{i,t} \min \Big( \rho_t^{(i)} A(a_t^{(i)}), \\
&\quad \text{clip}(\rho_t^{(i)}, 1\pm\epsilon) A(a_t^{(i)}) \Big)
\end{split}
\end{equation}
where $\rho_t^{(i)} = \pi_\theta(a_t^{(i)} \mid x_t^{(i)}) / \pi_\text{old}(a_t^{(i)} \mid x_t^{(i)})$ and $\omega$ balances the two advantage terms.

\section{Methodology}
\label{sec:method}

We present \methodname{} (Figure~\ref{fig:overall_framework}), a three-phase curriculum framework that turns skills from external runtime memories into temporary training-time signals. 
As motivated in Section~\ref{sec:intro}, \methodname{} aims to answer two questions: how to discover useful skills without external skill generators, and how to absorb these skills into a policy that no longer requires retrieval at inference time. 
The framework proceeds through: (1) policy warmup, (2) self-skill mining and utilization, and (3) advantage-weighted skill internalization.

For a task group $g$, we sample a group of trajectories $\mathcal{G}_g=\{\tau_i\}_{i=1}^N$. 
Each trajectory is assigned an arm indicator $z_i\in\{0,1\}$, where $z_i=1$ denotes the skill-augmented arm and $z_i=0$ denotes the skill-free baseline arm. 
Let $\mathcal{B}$ denote the skill bank, and let $\mathcal{M}_g=\mathcal{R}(\mathcal{B},q_g)$ be the retrieved skill set for task query $q_g$. 
We define the skill-free and skill-augmented contexts as
\begin{equation}
\label{eq:context_free}
x_t^{(i,-)}=[I_{\text{task}};\mathcal{H}_t^{(i)};o_t^{(i)}],
\end{equation}
and
\begin{equation}
\label{eq:context_skill}
x_t^{(i,+)}=[I_{\text{task}};\mathcal{M}_g;\mathcal{H}_t^{(i)};o_t^{(i)}],
\end{equation}
where $I_{\text{task}}$ is the task instruction, $o_t^{(i)}$ is the current observation, and $\mathcal{H}_t^{(i)}$ is the interaction history.
The policy acts under $x_t^{(i,+)}$ when $z_i=1$ and under $x_t^{(i,-)}$ when $z_i=0$.

\subsection{Phase 0: Policy Warmup}

Phase 0 bootstraps the agent's basic interaction ability and collects successful skill-free trajectories for later skill extraction. 
Directly mining skills from an untrained policy can introduce spurious heuristics, since low-quality trajectories may contain accidental or invalid behaviors. 
We therefore first train the policy with GiGPO~\citep{feng2025group}, which provides both episode-level and anchor-state step-level advantages without using an external critic.

Let $A^E(\tau_i)$ and $A^S(a_t^{(i)})$ denote the episode-level and step-level advantages computed by GiGPO.
The composite advantage used during warmup is
\begin{equation}
\label{eq:composite_adv}
A(a_t^{(i)}) = A^E(\tau_i) + \omega A^S(a_t^{(i)}),
\end{equation}
where $\omega$ balances trajectory-level and step-level credit.
The policy is optimized with the standard GiGPO objective $\mathcal{L}_{\text{GiGPO}}$.

\noindent \textbf{Phase transition.}
Let $k$ be the current training iteration, and let $N_{\text{succ}}$ count trajectories whose return exceeds a success threshold $\epsilon_{\text{succ}}$.
The warmup phase terminates once sufficient successful experience has been collected, or when a maximum warmup budget is reached:
\begin{equation}
\label{eq:warmup_exit}
\begin{split}
\mathbb{I}_{\text{warmup}} =
\mathbb{I}\big[
&(k \ge K_{\text{warm}} \wedge N_{\text{succ}}\ge N_{\text{req}})\\
&\vee\; k \ge K_{\text{max}}
\big].
\end{split}
\end{equation}
This adaptive transition ensures that skill mining starts only after the policy has produced enough reliable behavioral evidence, while preventing indefinite warmup in difficult environments.

\subsection{Phase 1: Self-Skill Mining and Utilization}

Phase 1 builds and validates a dynamic skill bank without relying on external skill generators. 
The key principle is that skills should be mined only from the agent's own successful skill-free behavior, and should be treated as candidate hypotheses until they are empirically verified.

\noindent \textbf{Self-skill mining.}
At the beginning of Phase 1, and then every $I_{\text{mine}}$ training iterations, the agent mines skills from successful trajectories in the skill-free arm. 
Specifically, only trajectories satisfying $z_i=0$ and $R(\tau_i)\ge \epsilon_{\text{succ}}$ are used as evidence. 
A frozen snapshot of the current policy summarizes these trajectories into compact natural-language skills. 
Each skill is represented as a condition--strategy pair
\begin{equation}
\label{eq:skill_repr}
m_j=(c_j,s_j,u_j,n_j,\text{state}_j),
\end{equation}
where $c_j$ describes when the skill applies, $s_j$ describes the reusable strategy, $u_j$ is its online utility estimate, $n_j$ is its retrieval count, and $\text{state}_j\in\{\texttt{candidate},\texttt{active},\texttt{retired}\}$.
Newly mined skills are inserted into $\mathcal{B}$ as \texttt{candidate} skills.

\noindent \textbf{Paired skill validation.}
For each task group $g$, we perform paired rollouts with both skill-free and skill-augmented arms. 
The utility of retrieved skills is estimated by the treatment effect between the two arms:
\begin{equation}
\label{eq:delta_g}
\Delta_g =
\overline{R}_g^{\text{skill}} -
\overline{R}_g^{\text{base}}.
\end{equation}
Here,
\begin{equation}
\label{eq:return_mean}
\begin{split}
\overline{R}_g^{\text{skill}}
&=
\frac{1}{|\mathcal{G}_g^{\text{skill}}|}
\sum_{\tau_i\in \mathcal{G}_g^{\text{skill}}} R(\tau_i), \\
\overline{R}_g^{\text{base}}
&=
\frac{1}{|\mathcal{G}_g^{\text{base}}|}
\sum_{\tau_i\in \mathcal{G}_g^{\text{base}}} R(\tau_i).
\end{split}
\end{equation}
where $\mathcal{G}_g^{\text{skill}}=\{\tau_i:z_i=1\}$ and $\mathcal{G}_g^{\text{base}}=\{\tau_i:z_i=0\}$.
For every retrieved skill $m_j$, its utility is updated by an exponential moving average:
\begin{equation}
\label{eq:skill_ema}
u_j \leftarrow \alpha u_j + (1-\alpha)\Delta_g,
\end{equation}
where $\alpha$ is the EMA coefficient.
After $n_j\ge N_{\text{eval}}$ retrievals, a candidate skill is promoted to \texttt{active} if $u_j>\epsilon_u$; skills with persistently negative utility are retired.
This lifecycle prevents self-generated skills from being trusted by default and keeps the skill bank grounded in online performance.

\noindent \textbf{Phase transition.}
The system enters Phase 2 once the skill bank becomes sufficiently mature:
\begin{equation}
\label{eq:phase2_exit}
\mathbb{I}\left[
|\mathcal{B}^{\text{active}}|\ge K_{\text{mat}}
\;\wedge\;
\bar{h}\ge\rho_{\text{hit}}
\;\wedge\;
\bar{\zeta}\ge\rho_{\text{pos}}
\right],
\end{equation}
where $K_{\text{mat}}$ is the minimum number of active skills, $\bar{h}$ is the recent retrieval hit rate, and $\bar{\zeta}$ is the recent proportion of task groups with $\Delta_g>0$.
This ensures that internalization begins only after retrieved skills have demonstrated consistent positive utility.

\subsection{Phase 2: Advantage-Weighted Skill Internalization}

\begin{table*}[t]
  \centering
  \setlength{\tabcolsep}{6.4pt}
  \begin{tabular}{l ccccccc | cc}
    \toprule
    \multirow{2}{*}{\textbf{Method}} 
    & \multicolumn{7}{c|}{\textbf{ALFWorld}} 
    & \multicolumn{2}{c}{\textbf{WebShop}} \\
    & Pick & Look & Clean & Heat & Cool & Pick2 & All & Score & Success \\
    \midrule

    \rowcolor{gray!15} \multicolumn{10}{l}{\textit{API-based LLMs}} \\
    Gemini-3-Flash 
    & 0.964 & \underline{0.857} & 0.571 & 0.722 & \textbf{0.962} & \textbf{0.953} & 0.852 & 0.141 & 0.165 \\
    GLM-5.1
    & 0.913 & 0.818 & 0.852 & \textbf{0.938} & 0.667 & 0.667 & 0.797 & 0.156 & 0.125 \\
    \midrule

    \rowcolor{gray!15} \multicolumn{10}{l}{\textit{Base Model: Qwen2.5-7B-Instruct}} \\
    Origin 
    & 0.179 & 0.643 & 0.048 & 0.000 & 0.038 & 0.053 & 0.125 & 0.166 & 0.039 \\
    \midrule

    \rowcolor{gray!15} \multicolumn{10}{l}{\textit{Prompt-based Agentic or Memory-based Methods}} \\
    ReAct 
    & 0.485 & 0.354 & 0.343 & 0.132 & 0.182 & 0.176 & 0.312 & 0.462 & 0.195 \\
    Reflexion 
    & 0.620 & 0.416 & 0.449 & 0.309 & 0.363 & 0.238 & 0.427 & 0.581 & 0.288 \\
    \midrule

    \rowcolor{gray!15} \multicolumn{10}{l}{\textit{RL-based Methods}} \\
    PPO 
    & 0.923 & 0.640 & 0.925 & 0.895 & 0.803 & 0.688 & 0.804 & 0.814 & 0.687 \\
    RLOO 
    & 0.876 & 0.782 & 0.873 & 0.813 & 0.719 & 0.489 & 0.755 & 0.803 & 0.657 \\
    GRPO 
    & 0.908 & 0.661 & 0.893 & 0.747 & 0.725 & 0.647 & 0.776 & 0.793 & 0.661 \\
    GiGPO 
    & \underline{0.977} & 0.827 & \underline{0.988} & 0.837 & 0.893 & 0.792 & \underline{0.908} & 0.844 & \underline{0.728} \\
    \midrule

    \rowcolor{gray!15} \multicolumn{10}{l}{\textit{Memory-Augmented RL-based Methods}} \\
    MemRL 
    & 0.628 & 0.385 & 0.222 & 0.125 & 0.080 & 0.000 & 0.214 & 0.295 & 0.092 \\
    EvolveR 
    & 0.649 & 0.333 & 0.464 & 0.133 & 0.333 & 0.333 & 0.438 & 0.425 & 0.176 \\
    Mem0+GRPO 
    & 0.781 & 0.548 & 0.561 & 0.310 & 0.650 & 0.269 & 0.547 & 0.581 & 0.375 \\
    SimpleMem+GRPO 
    & 0.895 & 0.363 & 0.600 & 0.500 & 0.649 & 0.263 & 0.625 & 0.678 & 0.469 \\
    SkillRL 
    & \textbf{0.979} & 0.714 & 0.900 & \underline{0.900} & \underline{0.955} & 0.875 & 0.899 & \underline{0.852} & 0.727 \\
    \midrule
    \textbf{\methodname{}} 
    & 0.975 & \textbf{0.875} & \textbf{1.000} & 0.857 & 0.850 & \underline{0.895} & \textbf{0.930} & \textbf{0.899} & \textbf{0.813} \\
    \bottomrule
  \end{tabular}
  \caption{
    \textbf{Performance on ALFWorld and WebShop.}
    For ALFWorld, we report the success rate on each subtask and the overall average.
    For WebShop, we report both the average score and the success rate.
    The best and second-best results among all listed methods are highlighted in \textbf{bold} and \underline{underline}, respectively.
  }
  \label{tab:performance}
\end{table*}

Phase 2 removes the dependency on the skill bank by transferring useful skill-guided behaviors into the skill-free policy. 
The policy acts as both teacher and student under asymmetric contexts. 
The teacher trajectory is generated under the skill-augmented context $x_t^{(i,+)}$, while the student is optimized to reproduce selected teacher action tokens under the skill-free context $x_t^{(i,-)}$.

Let $a_t^{(i,+)}$ be the teacher action from a skill-augmented trajectory, and let $\{y_{t,\ell}^{(i,+)}\}_{\ell=1}^{L_t}$ be its action tokens.
The student log-probability of the $\ell$-th teacher action token is
\begin{equation}
\label{eq:student_logprob}
\log p^{\text{stu}}_{i,t,\ell}
=
\log \pi_\theta
\left(
y_{t,\ell}^{(i,+)}
\,\middle|\,
x_t^{(i,-)},\text{sg}(y_{t,<\ell}^{(i,+)})
\right),
\end{equation}
where $\text{sg}(\cdot)$ denotes the stop-gradient operator.

\noindent \textbf{Composite utility gate.}
To avoid internalizing noisy or unhelpful skill-guided behaviors, we only distill action tokens from skill-augmented trajectories whose retrieved skills improve the task-level return:
\begin{equation}
\label{eq:utility_gate}
G_{i,t,\ell}
=
\mathbf{1}[z_i=1]\cdot
\mathbf{1}[\Delta_g>0]\cdot
\mathbf{1}[\mathrm{is\_action}(t,\ell)].
\end{equation}
Here $\mathrm{is\_action}(t,\ell)$ selects tokens inside the executable action span, excluding reasoning traces or other natural-language text.

\noindent \textbf{Advantage-weighted internalization.}
Among beneficial skill-augmented trajectories, not every action contributes equally to task success. 
We therefore reuse GiGPO's hierarchical credit $A(a_t^{(i)}) = A^E(\tau_i) + \omega A^S(a_t^{(i)})$ to weight the distillation signal. 
The token-level weight is
\begin{equation}
\label{eq:token_weight}
\kappa_{i,t,\ell}
=
G_{i,t,\ell}
\cdot
\frac{\max(A(a_t^{(i)}),0)}
{\overline{A}_{+}+\epsilon},
\end{equation}
where $\overline{A}_{+}$ is the mean positive advantage of the selected tokens within the current batch and $\epsilon$ is a small constant.
This weighting gives higher priority to skill-guided actions that receive stronger RL credit, while suppressing non-action tokens and low-utility trajectories.

\noindent \textbf{Joint objective.}
The internalization loss is
\begin{equation}
\label{eq:int_loss}
\mathcal{L}_{\text{int}}(\theta)
=
-\frac{1}{Z}
\sum_{i,t,\ell}
\kappa_{i,t,\ell}
\log p^{\text{stu}}_{i,t,\ell},
\end{equation}
where $Z=\max\left(1, \sum_{i,t,\ell}\mathbf{1}[\kappa_{i,t,\ell}>0]\right)$ normalizes over selected action tokens to prevent division by zero.
The final objective combines GiGPO optimization, KL regularization, and skill internalization:
\begin{equation}
\label{eq:total_loss}
\mathcal{L}_{\text{total}}(\theta)
=
\mathcal{L}_{\text{GiGPO}}(\theta)
+
\beta\mathcal{L}_{\text{KL}}(\theta)
+
\lambda(k)\mathcal{L}_{\text{int}}(\theta),
\end{equation}
where $\mathcal{L}_{\text{GiGPO}}(\theta)$ is defined in Section~\ref{sec:preliminaries}, $\beta$ controls policy deviation from the reference model, and $\lambda(k)$ is linearly warmed up after Phase 2 starts.

After training, the skill bank is discarded. 
At inference time, the policy runs only with the original skill-free context $x_t^{(i,-)}$, requiring no retrieval service, external memory, or skill prompt.

\section{Experimental Results}

\subsection{Performance on Agentic Tasks}
\label{sec:long_horizon_results}
\begin{table}[h]
  \centering
  \renewcommand{\arraystretch}{1.15}

  \begin{tabular}{l | cc}
    \toprule
    \textbf{Method} & \textbf{Score} & \textbf{Success} \\
    \midrule
    w/o Phase 0  & 0.850 & 0.711  \\
    w/o Phase 2 (w/ Skill) & \underline{0.898} & \underline{0.805}  \\
    w/o Phase 2 (w/o Skill) & 0.852 & 0.719  \\
    w/o Phase 1 and 2 (GiGPO)  & 0.844 & 0.728 \\
    \midrule
    \textbf{\methodname} & \textbf{0.899} & \textbf{0.813} \\
    \bottomrule
  \end{tabular}
  \caption{\label{tab:ablation_main}
    Ablation study of \methodname{} on WebShop. We report the average score and success rate with Qwen2.5-7B-Instruct backbone.
  }
\end{table}

\noindent \textbf{Experimental setup.}
We evaluate \methodname{} on two representative long-horizon agentic benchmarks: ALFWorld~\cite{shridhar2021alfworld} and WebShop~\cite{yao2022webshop}.
ALFWorld requires agents to complete embodied household tasks across six subtasks, including Pick, Look, Clean, Heat, Cool, and Pick2.
WebShop evaluates web-based product search and purchase behavior, where we report both the average score and the success rate.
We use Qwen2.5-7B-Instruct as the base model and compare \methodname{} with four groups of baselines.
First, we include API-based LLMs, including Gemini-3-Flash and GLM-5.1.
Second, we compare with prompt-based or memory-based agentic methods, including ReAct~\cite{yao2023react} and Reflexion~\cite{shinn2023reflexion}.
Third, we include standard RL-based methods, including PPO~\cite{schulman2017proximal}, RLOO~\cite{ahmadian2024back}, GRPO~\cite{shao2024deepseekmath}, and GiGPO~\cite{feng2025group}.
Finally, we compare with memory-augmented RL-based methods, including MemRL~\cite{zhang2026memrl}, EvolveR~\cite{wu2025evolver}, Mem0+GRPO~\cite{chhikara2025mem0}, SimpleMem+GRPO~\cite{liu2026simplemem}, and SkillRL~\cite{xia2026skillrl}.
All results are summarized in Table~\ref{tab:performance}, with further implementation details provided in Appendix~\ref{sec:appendix}.

\begin{table*}[t]
  \centering
  \setlength{\tabcolsep}{8.5pt}
  \begin{tabular}{l ccccccc | cc}
    \toprule
    \multirow{2}{*}{\textbf{Method}} & \multicolumn{7}{c|}{\textbf{ALFWorld}} & \multicolumn{2}{c}{\textbf{WebShop}} \\
    & Pick & Clean & Cool & Look & Heat & Pick2 & All & Score & Success \\
    \midrule

    \rowcolor{gray!15} \multicolumn{10}{l}{Base Model: \textit{Qwen2.5-1.5B-Instruct}} \\
    Origin & 0.059 & 0.033 & 0.042 & 0.055 & 0.097 & 0.000 & 0.041 & 0.231 & 0.052 \\
    ReAct & 0.174 & 0.157 & 0.077 & 0.205 & 0.062 & 0.020 & 0.128 & 0.401 & 0.113 \\
    Reflexion & 0.353 & 0.217 & 0.194 & 0.222 & 0.136 & 0.037 & 0.218 & 0.558 & 0.219 \\
    PPO & 0.648 & 0.571 & 0.464 & 0.405 & 0.606 & 0.474 & 0.544 & 0.738 & 0.515 \\
    RLOO & 0.883 & 0.710 & 0.664 & 0.528 & 0.628 & 0.569 & 0.697 & 0.739 & 0.521 \\
    GRPO & 0.853 & 0.845 & 0.597 & 0.537 & 0.782 & 0.535 & 0.728 & 0.758 & 0.568 \\
    GiGPO & \textbf{0.944} & \underline{0.948} & \underline{0.798} & \underline{0.675} & \underline{0.944} & \textbf{0.764} & \underline{0.867} & \underline{0.831} & \underline{0.650} \\
    \midrule
    \textbf{\methodname{}} & \underline{0.897} & \textbf{1.000} & \textbf{0.870} & \textbf{0.833} & \textbf{0.947} & \underline{0.600} & \textbf{0.875} & \textbf{0.853} & \textbf{0.727} \\
    \bottomrule
  \end{tabular}
  \caption{
  \label{tab:performance_1_5b}
    \textbf{Performance on ALFWorld and WebShop with Qwen2.5-1.5B-Instruct.}
    For ALFWorld, we report the success rate on each subtask and the overall average.
    For WebShop, we report both the average score and the success rate.
    The best and second-best results are highlighted in \textbf{bold} and \underline{underline}, respectively.
  }
\end{table*}

\begin{table}[t]
  \centering
  \renewcommand{\arraystretch}{1.15}
  \begin{tabular}{l | c c}
    \toprule
    \textbf{Method} & \textbf{Score} & \textbf{Success} \\
    \midrule
    GRPO & \underline{0.758} & \underline{0.568} \\
    \textbf{\methodname~(GRPO)} & \textbf{0.779}$_{\uparrow\textbf{0.021}}$ & \textbf{0.648}$_{\uparrow\textbf{0.080}}$ \\
    \midrule
    GiGPO  & \underline{0.831} & \underline{0.650} \\
    \textbf{\methodname~(GiGPO)} & \textbf{0.853}$_{\uparrow\textbf{0.022}}$ & \textbf{0.727}$_{\uparrow\textbf{0.077}}$ \\
    \bottomrule
  \end{tabular}
  \caption{
    Impact of different base optimization algorithms (GRPO vs. GiGPO) on the performance of \methodname{} on WebShop. We report the average score and success rate with Qwen2.5-1.5B-Instruct backbone.
  }
  \label{tab:ablation_grpo}
\end{table}

\begin{figure}[t]
  \includegraphics[width=\columnwidth]{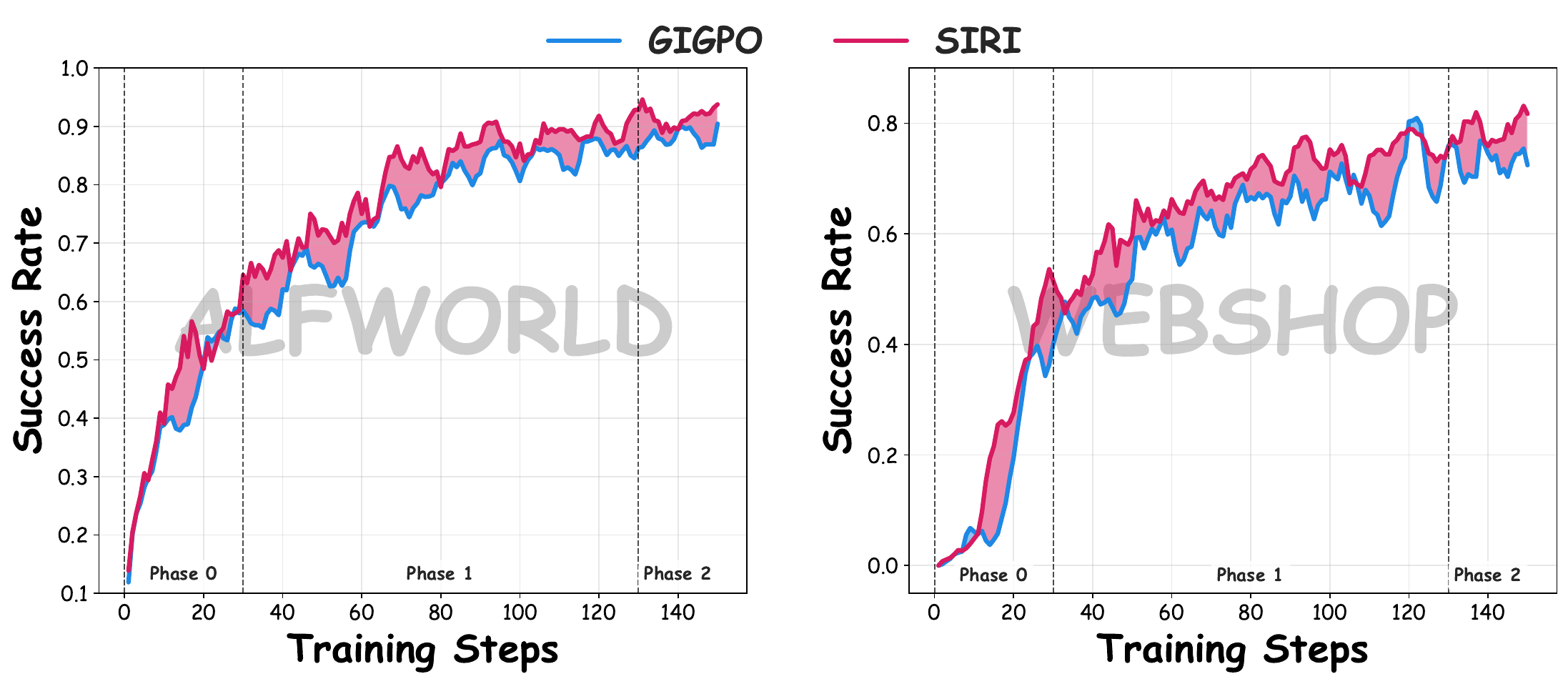}
  \caption{Training curves on ALFWorld and WebShop, showing that \methodname{} consistently improves over GiGPO throughout training.}
  \label{fig:compare_to_gigpo}
\end{figure}

\noindent \textbf{Experimental results.}
Table~\ref{tab:performance} reports the main results. We summarize three findings.
\noindent \textbf{1) Best overall performance.}
\methodname{} achieves the best overall results on both benchmarks.
On ALFWorld, it improves GiGPO from 0.908 to \textbf{0.930} and SkillRL from 0.899 to \textbf{0.930}.
On WebShop, it improves GiGPO from 0.728 to \textbf{0.813} in success rate and from 0.844 to \textbf{0.899} in score.
As shown in Figure~\ref{fig:compare_to_gigpo}, \methodname{} also maintains a stronger training trajectory than GiGPO, indicating that self-mined skills provide useful guidance during learning.
\noindent \textbf{2) Effective skill-free internalization.}
Compared with memory-augmented RL methods, \methodname{} achieves stronger final performance without relying on a runtime skill bank.
For example, SkillRL obtains 0.899 on ALFWorld and 0.727 on WebShop success, while \methodname{} reaches \textbf{0.930} and \textbf{0.813}.
This supports the central design of Advantage-Weighted Skill Internalization: useful skill-guided actions can be transferred into the model parameters, allowing the final policy to run with the original prompt only.
\noindent \textbf{3) Robust gains across tasks.}
\methodname{} achieves the best result on Look, Clean, ALFWorld overall, WebShop score, and WebShop success, and obtains the second-best result on Pick2.
These gains suggest that Self-Skill Mining and Utilization extracts reusable strategies from successful plain rollouts, while Advantage-Weighted Skill Internalization selectively absorbs high-utility action tokens.
Together, the two phases improve both embodied household interaction and web-based decision making under a unified skill-free inference framework.

\subsection{Ablation Study} 
Table \ref{tab:ablation_main} presents the ablation study of the \methodname{} framework on the WebShop benchmark. Removing the Phase 0 warmup (w/o Phase 0) incurs a substantial performance drop ($-10.2\%$), empirically demonstrating that initiating skill mining on a naïve policy inevitably injects severe noise into the skill repository. Furthermore, evaluating the Phase 1 checkpoint without retrieved skills (w/o Phase 2 (w/o Skill)) leads to a drastic degradation in success rate to $0.719$, revealing the agent's critical over-reliance on explicit privileged information. Conversely, evaluating this checkpoint with retrieved skills but without internalization (w/o Phase 2 (w/ Skill)) boosts performance to $0.805$, corroborating the efficacy and high quality of the mined skill bank. Crucially, by adopting an advantage-weighted skill internalization mechanism, the full \methodname{} framework effectively filters out irrelevant tokens and suboptimal trajectories, successfully distilling a highly refined and pure policy. Consequently, \methodname{} achieves a peak success rate of $0.813$ while completely eliminating retrieval dependency, thereby maximizing task performance and unlocking the computational efficiency inherent in retrieval-free generation.

\begin{figure}[t]
  \includegraphics[width=\columnwidth]{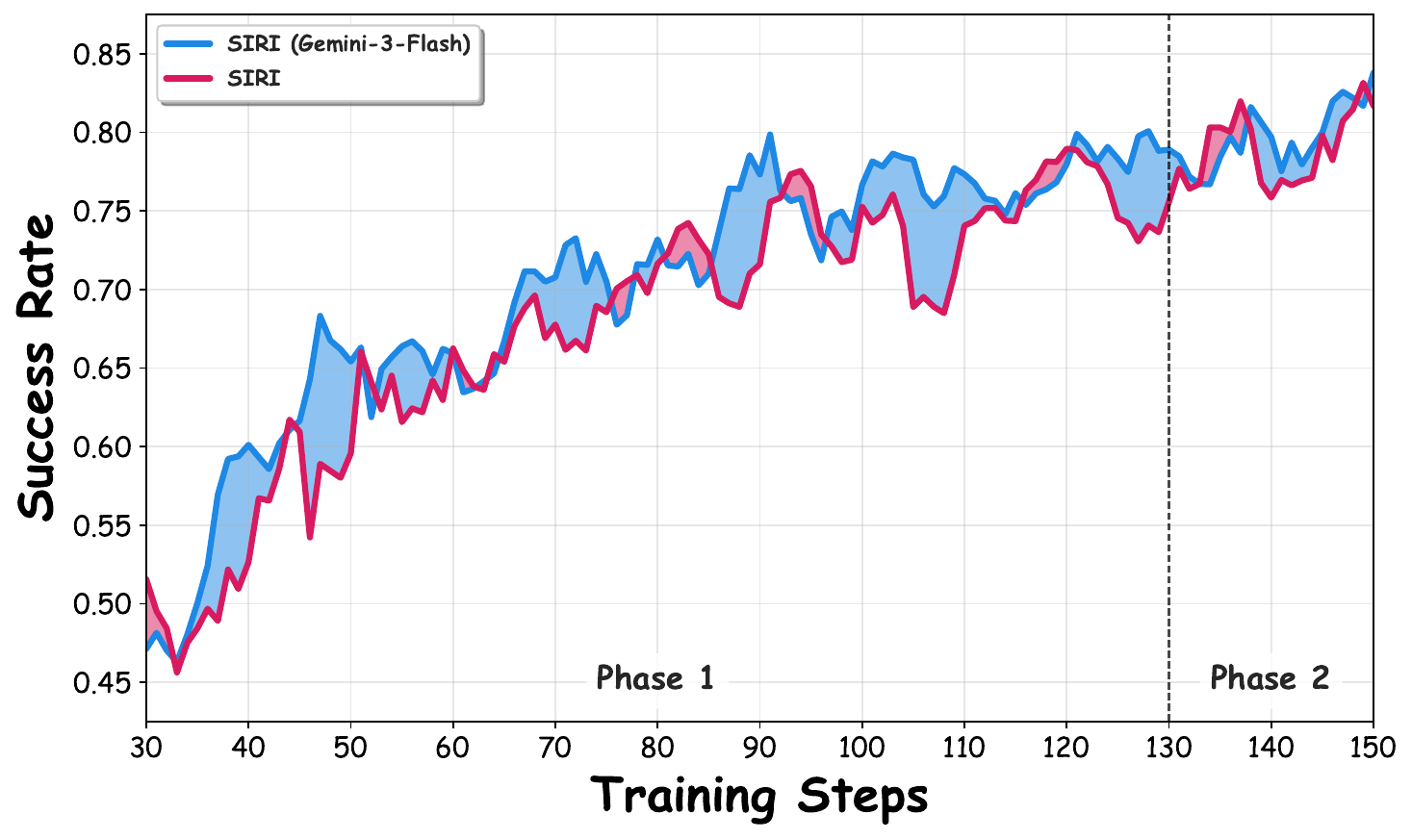}
  \caption{Training success rate trajectories on WebShop, comparing the standard \methodname{} against a variant utilizing Gemini-3-Flash for skill mining. }
  \label{fig:self_vs_external}
\end{figure}

\subsection{Results on Smaller Models}
\label{sec:more_comparisons}
To examine whether \methodname{} remains effective at smaller parameter scales, we further evaluate it with Qwen2.5-1.5B-Instruct as the base policy and report the results in Table~\ref{tab:performance_1_5b}. Despite the substantially reduced model capacity, \methodname{} still outperforms all baselines on both benchmarks: on ALFWorld, it improves the overall success rate of GiGPO from 0.867 to \textbf{0.875}, and on WebShop, it raises the success rate from 0.650 to \textbf{0.727} and the average score from 0.831 to \textbf{0.853}. These results indicate that the gains of \methodname{} do not rely on large model capacity: self-mined skills remain informative, and advantage-weighted internalization remains effective, even when the underlying policy is considerably smaller.

\subsection{Impact of Base Optimization Algorithms}
To further investigate the impact of the base reinforcement learning algorithm on the quality of skill mining and internalization, we compare \methodname{} instantiated on GRPO and GiGPO on WebShop, using Qwen2.5-1.5B-Instruct as the backbone, and report the results in Table~\ref{tab:ablation_grpo}. Two findings emerge. First, \methodname{} consistently improves both algorithms, raising the success rate from 0.568 to \textbf{0.648} on GRPO and from 0.650 to \textbf{0.727} on GiGPO, which shows that our framework is not tied to a specific policy optimizer. Second, \methodname{} built on GiGPO reaches a markedly higher ceiling than its GRPO counterpart. We attribute this gap to the difference in credit granularity: GRPO provides only episode-level signals, which limits the quality of policy warmup in long-horizon tasks, whereas GiGPO's anchor-state step-level advantages yield a stronger warmed-up policy and, consequently, cleaner successful trajectories for skill mining. Better behavioral evidence in turn produces a more refined skill bank, ultimately raising the upper bound of the internalized retrieval-free policy.

\subsection{Impact of External LLM Skill Mining} 
To investigate the impact of external LLM skill extraction on training dynamics and analyze the evolution of endogenous self-mined skills, we utilize Gemini-3-Flash for skill mining in Phase 1. Figure \ref{fig:self_vs_external} plots the training success rates of the standard \methodname{} alongside this Gemini-augmented variant. Initially, the variant exhibits a significantly steeper learning curve, rapidly establishing a performance advantage over the self-mining baseline. This highlights the initial quality gap between self-mined and oracle-level skills; by leveraging high-quality external skills, the policy effectively bypasses early exploration bottlenecks. However, as training progresses, this gap narrows substantially. While external skills accelerate early exploration, the quality of \methodname{}'s self-mined skills organically improves alongside its advancing capabilities. By increasingly leveraging experiences summarized from its own historical trajectories, the model gradually closes the performance gap. Ultimately, after Phase 2 internalization, the variant utilizing self-mined skills successfully achieves comparable performance to the one utilizing external LLM extraction.

\section{Conclusion}
\label{sec:conclusion}

We presented \methodname{}, a self-internalizing reinforcement learning framework for long-horizon LLM agents.
\methodname{} mines skills from the agent's own successful plain rollouts, validates them through online paired-rollout utility, and internalizes useful skill-guided actions via advantage-weighted distillation.
As a result, skills serve only as temporary training-time guidance, while the final policy requires no skill bank or retrieval service at inference time.
Experiments on ALFWorld and WebShop show that \methodname{} consistently outperforms standard RL baselines and memory-augmented agent methods, demonstrating that self-generated skills can be effectively absorbed into a retrieval-free policy.

\section*{Limitations}

Although \methodname\ establishes an effective self-evolving loop of skill discovery, utilization, and internalization, the agent's capability to mine high-quality skills currently scales implicitly with its overall proficiency. In the current framework, while the skill mining quality naturally advances alongside policy improvement, we do not introduce an explicit, dedicated training signal specifically tailored to optimize the self-mining process itself. For future work, we plan to explore the integration of explicit learning signals to further strengthen and systematically enhance the model's capacity for autonomous skill summarization.

\bibliography{camera_ready}

\appendix

\section{Implementation Details}
\label{sec:appendix}

We conduct our experiments on 8 A100 (80G) GPUs with the random seed set to 0. For the retrieval module, we utilize Qwen3-0.6B-Embedding as the embedding model, targeting the \texttt{when\_to\_apply} field for skill matching. Each training batch samples 16 tasks with a group size of 8, and a validation subset of 128 tasks is maintained; skill mining and model validation are performed every 10 steps, with the generation temperature set to 0.4 during validation. For the phase-transition criteria, the warmup phase lasts between $K_{\mathrm{warm}}=30$ and $K_{\mathrm{max}}=50$ steps and terminates once $N_{\mathrm{req}}=100$ successful trajectories are collected, while internalization begins once the skill bank contains at least $K_{\mathrm{mat}}=5$ active skills with a retrieval hit rate of $\rho_{\mathrm{hit}}=0.2$ and a fraction $\rho_{\mathrm{pos}}=0.1$ of task groups yielding positive utility. Paired rollouts are split evenly between the skill-augmented and skill-free arms (A/B ratio of 0.5), and the internalization loss coefficient $\lambda$ is set to 0.2; the same configuration is used across all benchmarks. For training schedules, ALFWorld runs for at most 50 steps with a prompt length limit of 3,072 tokens, and WebShop runs for at most 15 steps with a limit of 4,096 tokens; the maximum response length is constrained to 512 tokens for all tasks. In terms of computational overhead, training a 7B model requires approximately 39 hours on ALFWorld and 18 hours on WebShop, while a 1.5B model takes about 11 hours and 9.5 hours, respectively.

\section{Evaluating the Intrinsic Quality of Skills}

\begin{figure}[t]
  \includegraphics[width=\columnwidth]{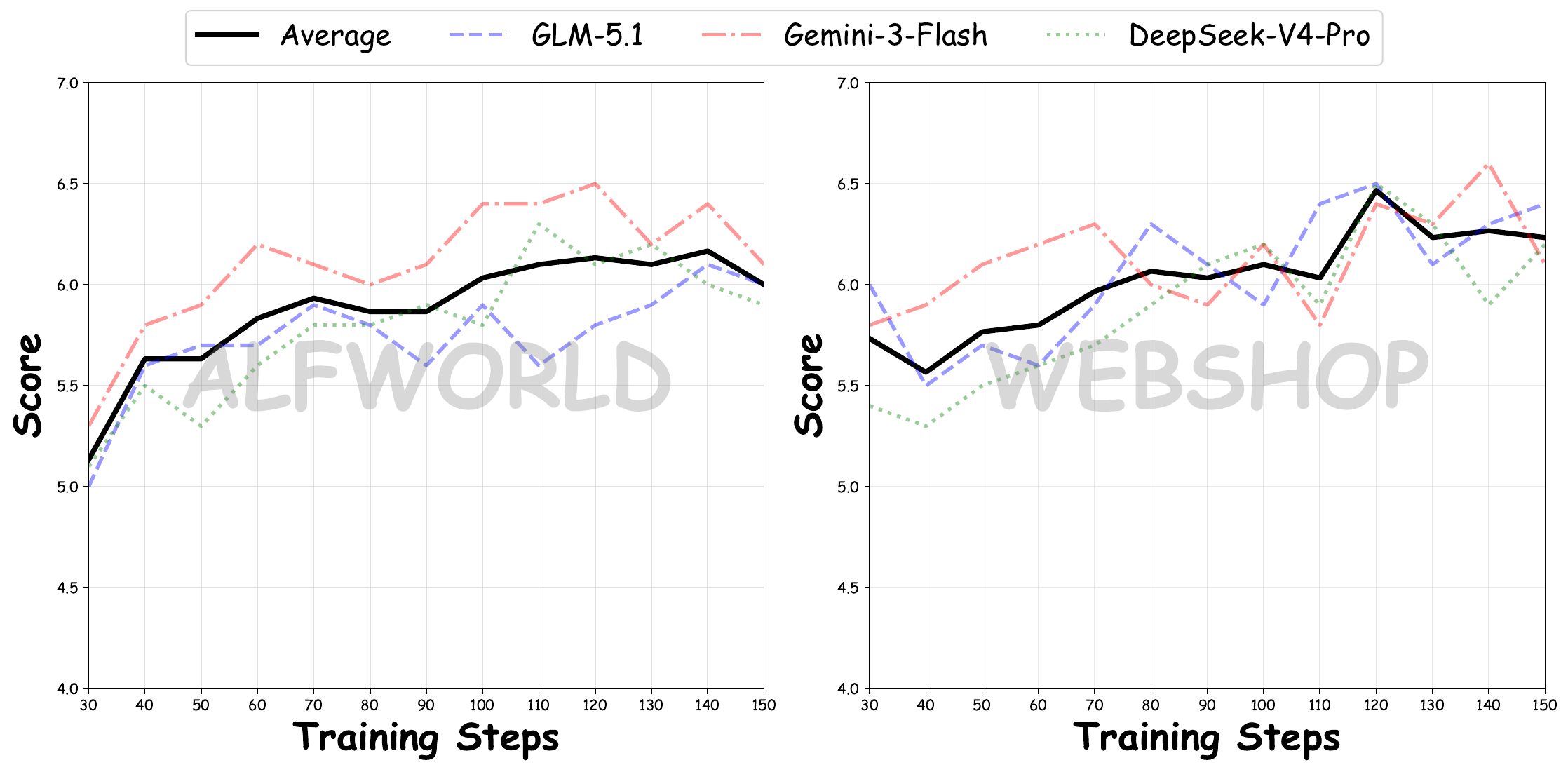}
  \caption{LLM-as-a-Judge quality scores for skills stored during training on ALFWorld and WebShop. The solid black line highlights the aggregated mean score across all evaluators, indicating continuous skill improvement.}
\label{fig:llm_judge}
\end{figure}

To evaluate the intrinsic quality of the skills generated during Phase 1, we employ a multi-model LLM-as-a-Judge framework (comprising Gemini-3-Flash, GLM-5.1, and DeepSeek-V4-Pro) to dynamically score the skills extracted throughout the training process, as illustrated in Figure~\ref{fig:llm_judge}. Across both the ALFWorld and WebShop benchmarks, while individual evaluator models exhibit local scoring variances, the aggregated mean score establishes a clear and sustained upward trend. This strong multi-model consensus indicates that as training progresses, the strategic skills mined by \methodname{} undergo substantial optimization in logical rigor and generalizability. Crucially, this steady enhancement in skill quality tightly correlates with the climbing task success rates observed in our main experiments. This alignment provides compelling evidence that the performance leaps of \methodname{} stem from the construction of a high-quality skill repository, rather than mere overfitting to specific environmental trajectories.

\section{Examples of Extracted Skills}
\label{appendix:skill_examples}

To provide a clearer understanding of the skills generated by our framework, we showcase concrete examples of extracted skills from both the ALFWorld and WebShop environments. Each skill abstracts a successful trajectory into a highly reusable heuristic, encompassing applicability conditions, a high-level strategy, and actionable key steps.

\noindent \textbf{ALFWorld.} 
The following skill demonstrates how the agent learns to overcome exploration bottlenecks when a target object is absent from its initial expected location.

\begin{tcolorbox}[
    colback=gray!4,
    colframe=gray!45,
    boxrule=0.5pt,
    arc=2mm,
    left=1.5mm,
    right=1.5mm,
    top=1mm,
    bottom=1mm,
    breakable,
    title=\textbf{Skill Example},
    fonttitle=\small,
    coltitle=black,
    colbacktitle=gray!12
]
\small
\textbf{When to apply:} When the required object is not found at the initial search location and must be located to proceed with the task. 

\textbf{Strategy:} Evaluate the current observation to confirm the object's absence, then systematically navigate to the next most logical storage receptacle based on the object's typical context. 

\textbf{Key steps:} Observe the current location | Verify the target object is absent | Identify the next probable storage receptacle | Navigate to the new location
\end{tcolorbox}

\noindent \textbf{WebShop.} 
The following skill illustrates how the agent systematically configures product attributes and verifies task constraints prior to executing a purchase.

\begin{tcolorbox}[
    colback=gray!4,
    colframe=gray!45,
    boxrule=0.5pt,
    arc=2mm,
    left=1.5mm,
    right=1.5mm,
    top=1mm,
    bottom=1mm,
    breakable,
    title=\textbf{Skill Example},
    fonttitle=\small,
    coltitle=black,
    colbacktitle=gray!12
]
\small
\textbf{When to apply:} When a product page is open and specific attribute variations must be chosen to match task requirements. 

\textbf{Strategy:} Configure the product by selecting all required options, confirm that constraints like price remain satisfied, and finalize the transaction. 

\textbf{Key steps:} Select required item attributes | Verify constraints are met | Execute the purchase action
\end{tcolorbox}

\section{Prompt Templates}
\label{app:prompt_templates}

\noindent \textbf{ALFWorld prompt.}
For ALFWorld, we use the following prompt template with skill memory augmentation.

\begin{tcolorbox}[
    colback=gray!4,
    colframe=gray!45,
    boxrule=0.5pt,
    arc=2mm,
    left=1.5mm,
    right=1.5mm,
    top=1mm,
    bottom=1mm,
    breakable,
    title=\textbf{ALFWorld Prompt Template (with Skill Memory)},
    fonttitle=\small,
    coltitle=black,
    colbacktitle=gray!12
]
\small

You are an expert decision-making agent operating in the ALFRED embodied environment.

Your goal is to complete the following task: \{task\_description\}

\textbf{Current Progress}

\smallskip
You have already taken \{step\_count\} step(s).

Recent interaction history (observation $\rightarrow$ action): \{action\_history\}

Current step: \{current\_step\}

Current observation: \{current\_observation\}

Admissible actions at this step: [\{admissible\_actions\}]

\textbf{Relevant Experience}

\smallskip
Below are past experiences retrieved from memory. Each experience is formatted as:

\hspace{1em}\textbullet\ \textbf{When to apply}: the condition under which this skill is relevant.

\hspace{1em}\textbullet\ \textbf{Strategy}: the overall approach that worked for this kind of task.

\hspace{1em}\textbullet\ \textbf{Key steps}: a concrete sequence of actions to follow, separated by ``|''.

\smallskip
When reasoning, you may:

\hspace{1em}\textbullet\ Check the bold condition first --- only use an experience if your current situation matches it.

\hspace{1em}\textbullet\ Use the strategy to guide your overall plan and avoid known pitfalls.

\hspace{1em}\textbullet\ Use the steps as a reference action sequence when the task situation is similar.

\hspace{1em}\textbullet\ Adapt the steps to your current observation rather than following them blindly.

\smallskip
\textit{Warning: These lessons may be outdated. Use them only if they align with your current observation.}

\smallskip
Retrieved experiences: \{retrieved\_memories\}

\textbf{Instructions}

\smallskip
For the current step, you should follow this process:

\hspace{1em}1.\ Analyze the current observation.

\hspace{1em}2.\ Review the retrieved experiences and think about whether any past experience applies.

\hspace{1em}3.\ Reason step-by-step and choose the best admissible action.

Now it's your turn to take an action. You should first reason step-by-step about the current situation. This reasoning process MUST be enclosed within \texttt{<think>} \texttt{</think>} tags.

\smallskip
Once you've finished your reasoning, you should choose an admissible action for current step and present it within \texttt{<action>} \texttt{</action>} tags.
\end{tcolorbox}

\noindent \textbf{WebShop prompt.}
For WebShop, we use the following prompt template with skill memory augmentation.

\begin{tcolorbox}[
    colback=gray!4,
    colframe=gray!45,
    boxrule=0.5pt,
    arc=2mm,
    left=1.5mm,
    right=1.5mm,
    top=1mm,
    bottom=1mm,
    breakable,
    title=\textbf{WebShop Prompt Template (with Skill Memory)},
    fonttitle=\small,
    coltitle=black,
    colbacktitle=gray!12
]
\small

You are an expert autonomous agent operating in the WebShop e-commerce environment.

Your goal is to complete the following task: \{task\_description\}

\textbf{Current Progress}

\smallskip
You have already taken \{step\_count\} step(s).

Recent interaction history (observation $\rightarrow$ action): \{action\_history\}

Current step: \{current\_step\}

Current observation: \{current\_observation\}

Admissible actions at this step: [\{available\_actions\}]

\textbf{Relevant Experience}

\smallskip
Below are past experiences retrieved from memory. Each experience is formatted as:

\hspace{1em}\textbullet\ \textbf{When to apply}: the condition under which this skill is relevant.

\hspace{1em}\textbullet\ \textbf{Strategy}: the overall approach that worked for this kind of task.

\hspace{1em}\textbullet\ \textbf{Key steps}: a concrete sequence of actions to follow, separated by ``|''.

\smallskip
When reasoning, you may:

\hspace{1em}\textbullet\ Check the bold condition first --- only use an experience if your current situation matches it.

\hspace{1em}\textbullet\ Use the strategy to guide your overall plan and avoid known pitfalls.

\hspace{1em}\textbullet\ Use the steps as a reference action sequence when the task situation is similar.

\hspace{1em}\textbullet\ Adapt the steps to your current observation rather than following them blindly.

\smallskip
\textit{Warning: These lessons may be outdated. Use them only if they align with your current observation.}

\smallskip
Retrieved experiences: \{retrieved\_memories\}

\textbf{Instructions}

\smallskip
For the current step, you should follow this process:

\hspace{1em}1.\ Analyze the current observation.

\hspace{1em}2.\ Review the retrieved experiences and think about whether any past experience applies.

\hspace{1em}3.\ Reason step-by-step and choose the best admissible action.

Now it's your turn to take an action. You should first reason step-by-step about the current situation. This reasoning process MUST be enclosed within \texttt{<think>} \texttt{</think>} tags.

\smallskip
Once you've finished your reasoning, you should choose an admissible action for current step and present it within \texttt{<action>} \texttt{</action>} tags.
\end{tcolorbox}

\noindent \textbf{Skill mining prompt.}
To extract reusable strategies from successful trajectories, we use the following template to summarize behavioral patterns into structured skill representations. The model is required to output exactly three labeled fields in plain text.

\begin{tcolorbox}[
    colback=gray!4,
    colframe=gray!45,
    boxrule=0.5pt,
    arc=2mm,
    left=1.5mm,
    right=1.5mm,
    top=1mm,
    bottom=1mm,
    breakable,
    title=\textbf{Skill Mining Prompt Template},
    fonttitle=\small,
    coltitle=black,
    colbacktitle=gray!12
]
\small

You are extracting a reusable task strategy from a successful agent trajectory. Output EXACTLY three labeled lines. No markdown. No bullets. No extra text.

\textbf{=== FORMAT EXAMPLE 1 ===}

\smallskip
Task: [example task A]

Action sequence:

\hspace{1em}Step 1: locate the target object

\hspace{1em}Step 2: verify it matches the required attributes

\hspace{1em}Step 3: execute the required action on it

\hspace{1em}Step 4: confirm the result

\smallskip
When to apply: When the task requires locating a specific object and performing a targeted action on it.

Strategy: First verify the object matches all required attributes before acting, to avoid wasted steps.

Key steps: Locate the target object | Verify it matches all required attributes | Execute the action | Confirm the result

\textbf{=== FORMAT EXAMPLE 2 ===}

\smallskip
Task: [example task B]

Action sequence:

\hspace{1em}Step 1: attempt action with initial parameters

\hspace{1em}Step 2: observe the outcome

\hspace{1em}Step 3: adjust parameters based on outcome

\hspace{1em}Step 4: retry and complete

\smallskip
When to apply: When an initial attempt fails and the task requires iterative refinement to succeed.

Strategy: After a failed attempt, diagnose the mismatch and adjust the key parameter before retrying.

Key steps: Attempt with initial parameters | Observe the outcome | Adjust the parameter causing failure | Retry until success

\textbf{=== YOUR TASK ===}

\smallskip
Task: \{task\}

Action sequence (completed successfully): \{actions\}

\textbf{Rules:}

\hspace{1em}\textbullet\ Output EXACTLY three lines with these labels: ``When to apply:'', ``Strategy:'', ``Key steps:''.

\hspace{1em}\textbullet\ Key steps: 2--4 steps separated by ``|'' (pipe with spaces). No numbers, no bullets.

\hspace{1em}\textbullet\ Do NOT mention specific object names, locations, product IDs, or step numbers.

\hspace{1em}\textbullet\ Focus on generalizable patterns applicable to similar tasks.

\hspace{1em}\textbullet\ Keep each field under 2 sentences.

When to apply:
\end{tcolorbox}

\section{Pseudo Code of \methodname{}}
\label{app:pseudo_code}

Algorithm~\ref{alg:method} presents the training procedure of \methodname{}.

\begin{algorithm*}[t]
    \small
    \caption{\methodname{} Training Procedure}
    \label{alg:method}
    \begin{algorithmic}[1]
        \STATE \textbf{Input:} training task set $\mathcal{D}$,
        policy $\pi_\theta$,
        reference policy $\pi_{\mathrm{ref}}$,
        environment $\mathcal{E}$,
        group size $N$,
        maximum iterations $K$,
        warmup thresholds $K_{\mathrm{warm}},K_{\max},N_{\mathrm{req}}$,
        success threshold $\epsilon_{\mathrm{succ}}$,
        mining interval $I_{\mathrm{mine}}$,
        skill maturity thresholds $K_{\mathrm{mat}},\rho_{\mathrm{hit}},\rho_{\mathrm{pos}}$,
        step-credit weight $\omega$,
        KL coefficient $\beta$,
        internalization schedule $\lambda(k)$.
        
        \STATE \textbf{Initialize:} skill bank $\mathcal{B}\leftarrow\emptyset$,
        phase indicator $p\leftarrow 0$,
        success counter $N_{\mathrm{succ}}\leftarrow 0$.
        
        \STATE \parbox[t]{\linewidth}{\centering \textit{*** \methodname{} training begins ***}}
        
        \FOR{$k=1$ to $K$}
            \STATE Set old policy $\pi_{\theta_{\mathrm{old}}}\leftarrow \pi_\theta$.
            
            \FOR{each task group $g\in\mathcal{D}$}
                \STATE \parbox[t]{\linewidth}{\centering \textit{*** Step A: Phase-aware rollout collection. ***}}
                
                \IF{$p=0$}
                    \STATE Sample $N$ skill-free trajectories under $x_t^{(i,-)}$ in Eq.~\eqref{eq:context_free}.
                \ELSE
                    \STATE Retrieve skills $\mathcal{M}_g=\mathcal{R}(\mathcal{B},q_g)$ and sample paired skill-free / skill-augmented trajectories under Eqs.~\eqref{eq:context_free}--\eqref{eq:context_skill}.
                \ENDIF
                
                \STATE Compute trajectory returns $\{R(\tau_i)\}_{i=1}^{N}$ and update $N_{\mathrm{succ}}$.
                
                \STATE \parbox[t]{\linewidth}{\centering \textit{*** Step B: GiGPO credit computation. ***}}
                
                \STATE Compute episode-level advantages $A^E(\tau_i)$ and step-level advantages $A^S(a_t^{(i)})$ using GiGPO.
                \STATE Compute composite action advantages $A(a_t^{(i)})=A^E(\tau_i)+\omega A^S(a_t^{(i)})$.
                
                \STATE \parbox[t]{\linewidth}{\centering \textit{*** Step C: Self-skill mining and validation. ***}}
                
                \IF{$p\ge 1$}
                    \IF{$k \bmod I_{\mathrm{mine}}=0$}
                        \STATE Mine candidate skills from successful skill-free trajectories with $R(\tau_i)\ge\epsilon_{\mathrm{succ}}$.
                        \STATE Insert mined skills $m_j=(c_j,s_j,u_j,n_j,\mathrm{state}_j)$ into $\mathcal{B}$ as \texttt{candidate}.
                    \ENDIF
                    
                    \STATE Compute paired-rollout utility $\Delta_g$ by Eqs.~\eqref{eq:delta_g}--\eqref{eq:return_mean}.
                    \STATE Update retrieved skill utilities by Eq.~\eqref{eq:skill_ema}.
                    \STATE Promote positive-utility candidate skills to \texttt{active} and retire persistently negative skills.
                \ENDIF
                
                \STATE \parbox[t]{\linewidth}{\centering \textit{*** Step D: Advantage-weighted internalization. ***}}
                
                \IF{$p=2$}
                    \STATE Compute student log-probabilities under skill-free contexts by Eq.~\eqref{eq:student_logprob}.
                    \STATE Construct utility gates $G_{i,t,\ell}$ by Eq.~\eqref{eq:utility_gate}.
                    \STATE Compute token weights $\kappa_{i,t,\ell}$ by Eq.~\eqref{eq:token_weight}.
                    \STATE Compute internalization loss $\mathcal{L}_{\mathrm{int}}$ by Eq.~\eqref{eq:int_loss}.
                    \STATE Optimize $\pi_\theta$ with the joint objective in Eq.~\eqref{eq:total_loss}.
                \ELSE
                    \STATE Optimize $\pi_\theta$ with $\mathcal{L}_{\mathrm{GiGPO}}+\beta\mathcal{L}_{\mathrm{KL}}$.
                \ENDIF
                
            \ENDFOR
            
            \STATE \parbox[t]{\linewidth}{\centering \textit{*** Step E: Phase transition. ***}}
            
            \IF{$p=0$ and Eq.~\eqref{eq:warmup_exit} is satisfied}
                \STATE Set $p\leftarrow 1$.
            \ENDIF
            
            \IF{$p=1$ and Eq.~\eqref{eq:phase2_exit} is satisfied}
                \STATE Set $p\leftarrow 2$.
            \ENDIF
            
        \ENDFOR
        
        \STATE Discard skill bank $\mathcal{B}$.
        \STATE \textbf{Output:} retrieval-free policy $\pi_\theta$ using only the skill-free context $x_t^{(i,-)}$.
    \end{algorithmic}
\end{algorithm*}

\end{document}